\documentclass[lettersize,journal]{IEEEtran}
\usepackage{amsmath,amsfonts}
\usepackage{algorithm}
\usepackage{array}
\usepackage[caption=false,font=normalsize,labelfont=sf,textfont=sf]{subfig}
\usepackage{textcomp}
\usepackage{stfloats}
\usepackage{url}
\usepackage{verbatim}
\usepackage{graphicx}
\usepackage{cite}
\usepackage{cite}
\usepackage{booktabs}
\usepackage{multirow}
\usepackage{graphicx} 
\usepackage{hyperref}
\usepackage{makecell}
\usepackage{mathtools}
\usepackage{tabularx}
\usepackage{algpseudocode} 

\usepackage{color}
\usepackage[table]{xcolor} 

\begin{document}

\title{\huge RSGPNet: Geometric Prompting for Remote Sensing Open-Vocabulary Semantic Segmentation}

\author{\IEEEauthorblockN{Shanwen Wang, Xin Sun,~\IEEEmembership{Senior member,~IEEE}, Sirui Wang,  Xiao Xiang Zhu,~\IEEEmembership{Fellow,~IEEE} 
}
\thanks{This work is supported by the Science and Technology Development Fund - International Collaborative Research, Macao SAR (0001/2025/AIJ), Science and Technology Development Fund, Macao SAR - Ministry of Science and Technology: National Key R\&D Program of China (0007/2025/AMJ, 2025YFE0202900), and Science and Technology Development Fund, Macao SAR - Basic Research (0006/2024/RIA1)} 
\thanks{S.W. Wang and X. Sun are with Faculty of Data Science, City University of Macau, 999078, SAR Macao, China. (Corresponding email: sunxin1984@ieee.org)}
\thanks{S.R. Wang and X.X. Zhu are with the Chair of Data Science in Earth Observation, Technical University of Munich (TUM), Germany and also with the Munich Center for Machine Learning, Munich, Germany.}
}

\markboth{Journal of \LaTeX\ Class Files,~Vol.~14, No.~8, August~2021}%
{Shell \MakeLowercase{\textit{et al.}}: A Sample Article Using IEEEtran.cls for IEEE Journals}

\maketitle

\begin{abstract}
Open-vocabulary semantic segmentation (OVSS) enables text-guided segmentation of unseen objects, breaking fixed-class limitations to achieve open-world understanding. However, existing OVSS methods primarily focus on modifying the CLIP attention mechanism, which still suffers from unstable local segmentation for remote sensing (RS) domain. To address these limitations, we propose RSGPNet, a training-free geometric prompting framework for RS OVSS that refines segmentation by leveraging object geometric areas and consistency constraints. Specifically, RSGPNet comprises three core modules: a Text-guided Coarse Mask module (TCM), a Geometric Re-prompting Module (GRP), and a Coarse-to-fine Consistency Verification Mechanism (CVM). TCM utilizes text prompts and the input image to construct initial coarse segmentation masks. GRP then converts these coarse masks into geometric box prompts, feeding them back into the segmentation model to generate refined masks. Finally, CVM employs consistency computation to prevent prompting from reinforcing erroneous regions. They allow the model to improve segmentation accuracy in complex areas, such as category boundaries. Extensive experiments on RS datasets demonstrate that RSGPNet significantly outperforms state-of-the-art methods across both quantitative and qualitative metrics while exhibiting excellent interpretability. The code is released at 
\href{https://github.com/wangshanwen001/RSGPNet}{https://github.com/wangshanwen001/RSGPNet}.
\end{abstract}


\section{Introduction}
\IEEEPARstart{S}{emantic} segmentation is a fundamental computer vision task that assigns a semantic label to every pixel in an image\cite{li2026frequency, li2025enhanced}. It plays a crucial role in remote sensing (RS) by interpreting satellite and aerial imagery, enabling the automatic extraction of land-cover information for applications such as urban management, environmental monitoring, and disaster prevention \cite{wang2024sclip, 11062866}. However, most existing segmentation models rely on a closed-set assumption with predefined training categories. Consequently, they struggle to recognize and segment unseen categories in real-world scenarios\cite{wang2025dual, xu2023side}. To overcome this limitation, open-vocabulary semantic segmentation (OVSS) segment arbitrary objects described by natural language, thereby extending recognition capabilities beyond the closed set of training labels \cite{wang2026openurban3d, faulkenberry2026dino, zhang2026towards, li2026exploring, zhao2026open}. OVSS models typically utilize Contrastive Language–Image Pretraining (CLIP) to align visual and textual representations \cite{shao2024explore, li2025segearth, zhang2023simple, sun2024clip, luo2023segclip}.  By leveraging the rich semantic knowledge embedded in large-scale vision–language pretrained model, they can segment open-vocabulary categories without requiring category-specific supervision. However, current RS-based OVSS methods suffer from imprecise local segmentation. Because their cross-modal alignment is learned at the image level rather than the dense pixel level, they fail to accurately capture complex object boundaries. While existing frameworks like OVRS \cite{10962188}, RSCLIP \cite{11502025}, and SegEarth-OV \cite{li2025segearth} have partially addressed these domain-specific challenges, their overall effectiveness in RS applications is still heavily limited. This bottleneck arises because they rely on conventional strategies, such as multi-scale and self-attention mechanisms, to solve RS OVSS challenges. Specifically, while some approaches attempt to refine the CLIP attention mechanism, they remain fundamentally bottlenecked by the inherent constraints of the pre-trained CLIP model. Consequently, this strict reliance on an image-text matching paradigm prevents existing OVSS methods from achieving precise local region segmentation in complex RS images.

 \begin{figure}[!t]
\centering
\includegraphics[width=3.5in]{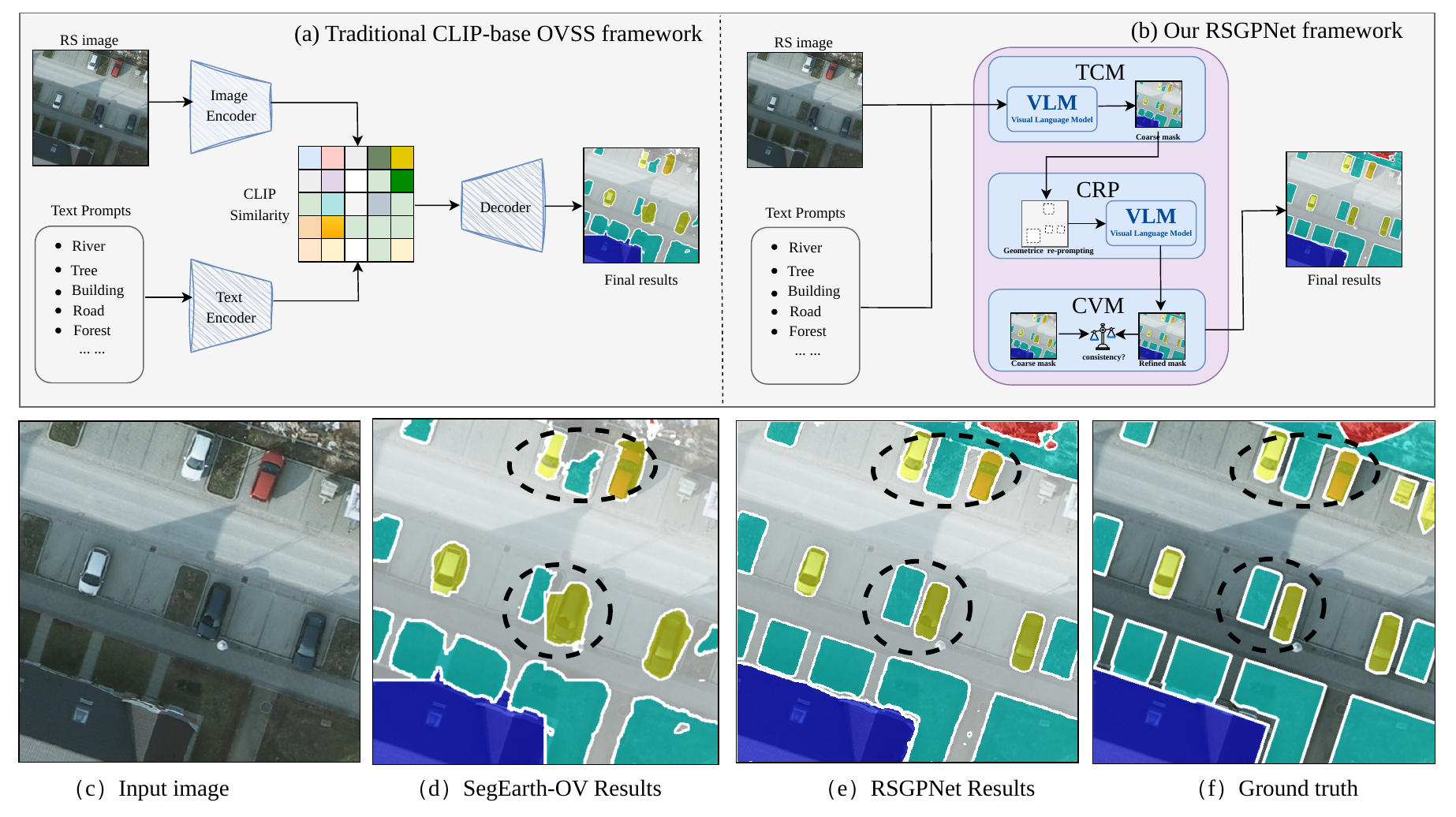}
\vspace{-0.5cm}
\caption{Comparison between the traditional CLIP-base OVSS framework (a) and ours RSGPNet (b). Traditional CLIP-based OVSS methods often produce imperfect segmentation results, whereas RSGPNet, empowered by geometric prompts, exhibits a stronger capability for fine-grained segmentation.}
\label{fig: fg1}
\vspace{-0.5cm}
\end{figure}

To address these challenges, we introduce RSGPNet, a novel, training-free OVSS framework tailored for RS. Departing from conventional CLIP-based architectures, our method establishes a new state-of-the-art (SOTA) in RS OVSS through a novel geometric prompting paradigm. Specifically, RSGPNet is driven by three key innovations. First, we propose the Text-guided Coarse Mask module (TCM), which leverages text prompts and the input image to construct initial coarse segmentation masks. Second, RSGPNet introduces a Geometric Re-prompting Module (GRP), which converts high-confidence regions into geometric box prompts and feeds them back into the segmentation model to generate refined masks. Finally, we introduce a Coarse-to-fine Consistency Verification Mechanism (CVM) to validate the fine-grained segmentation and prevent prompting from reinforcing erroneous regions. As illustrated in Fig. \ref{fig: fg1}, unlike conventional OVSS methods, our approach does not rely on the CLIP model and achieves coarse-to-fine fine-grained segmentation through geometric prompting. In particular, RSGPNet exhibits exceptional local-region segmentation performance across a wide range of categories. As highlighted by the black dashed ellipse in Fig. \ref{fig: fg1}(d), the CLIP-based SegEarth-OV exhibits noticeable artifacts in local detail segmentation due to the inherent shortcomings of CLIP. In contrast, RSGPNet achieves a visually discernible improvement in segmentation granularity, delivering superior accuracy and well-defined results.

In summary, our contributions are as follows:

\begin{enumerate}
\item We propose RSGPNet, a novel training-free framework for RS OVSS. We are the first to introduce geometric prompts to address the critical bottleneck in local-region segmentation caused by conventional image-text contrastive learning.
\item We introduce a coarse-to-fine mask generation paradigm with TCM and GRP to improve local region refinement. TCM first extracts an initial coarse mask, from which GRP derives geometric prompts that are fed back into the segmentation model for refinement.
\item We introduce the CVM to prevent geometric prompts from reinforcing incorrectly segmented regions. It effectively reduces discrepancies between the re-prompted segmentations and the originals, by evaluating the consistency between the coarse and refined masks.
\item Extensive experimental results demonstrate the efficiency of our RSGPNet model compared to SOTA methods. The results highlight the significance of our contributions to learning fine-grained features, thereby enhancing segmentation performance.
\end{enumerate}

The rest of this article is organized as follows: Section \ref{sec: related work} provides an overview of existing related research. In Section \ref{sec: method}, we formally propose and analyze our RSGPNet model. Section \ref{sec: experiment} presents comprehensive experimental results, including comparative analyses with SOTA methods and ablation studies. Finally, Section V concludes and discusses the article.

\section{related work}
\label{sec: related work}

This section reviews relevant research on OVSS, and also summarizes recent progress in OVSS for RS domain.

\subsection{Open-Vocabulary Semantic Segmentation}
OVSS aims to go beyond the limitations of traditional models constrained by closed-set categories, enabling the recognition and segmentation of unseen classes\cite{park2025understanding,lee2025effective, lai2025exploring,li2025towards,ge2025clip}. Current mainstream approaches predominantly leverage pre-trained vision-language models (VLMs), such as CLIP, to achieve zero-shot knowledge transfer by computing cross-modal similarity between image patches and text descriptions\cite{wang2026adaptive, wang2026rethinking,shao2026excluding, yuan2026infoclip, ye2026exploiting}. Representative works typically involve fine-tuning CLIP for segmentation tasks or employing strategies like prompt learning and adapter networks to generate mask proposals\cite{zemskova2026ovsegdt,pu2026panoearth,gong2026ov3r,chng2026aligning,shi2025llmformer}. Representative models include MaskCLIP\cite{dong2023maskclip}, SCLIP\cite{wang2024sclip}, ClearCLIP\cite{lan2024clearclip}, and others. These methods demonstrate that the semantic priors learned from large-scale image-text pairs can provide strong category-level generalization and improve segmentation performance under open-vocabulary settings. Moreover, recent studies further explore more effective visual-textual alignment, region-level representation learning, and prompt optimization to reduce the gap between image-level VLM pre-training and dense prediction tasks\cite{liang2023open, ge2025clip, wang2026openurban3d}. 

Although these methods have achieved remarkable progress on natural images, they often overlook the unique characteristics of RS imagery, such as complex multi-class local regions and significant inter-class scale variations. Therefore, a key challenge involves effectively transferring VLM knowledge to the RS domain and developing specialized alignment and fine-tuning mechanisms for RS data\cite{yang2026test, su2026seg2change, huang2026reducing}.

\subsection{Open-Vocabulary Semantic Segmentation for RS Images}
As discussed in the previous subsection, OVSS methods developed for the natural image domain cannot address the challenges in the RS domain. In the process of migrating from the OVSS field to the RS field, OVRS\cite{10962188} takes into account the basic characteristics of the RS field. Their methods calculated feature correlations between rotated image embeddings and text embeddings, leveraging an attention-aware upsampling decoder to generate segmentation outputs. GSNet\cite{ye2025towards} introduces a specialist-generalist model, which can better integrate RS domain knowledge into the model. Li et al.\cite{li2025segearth} proposed the SegEarth-OV model, it uses a global bias alleviation mechanism removed the influence of CLIP’s global attributes on local features. RSCLIP\cite{11502025} addresses the critical challenges of local region segmentation and significant object scale variations in RS images. To address the absence of a unified evaluation benchmark, Li et al.\cite{li2026exploring} constructed OVRSISBench, a benchmark for the RS OVSS domain. Ye et al.\cite{ye2026exploiting} proposed the LDSeg model to enhance the spatial discriminability and cross-modal attention of the CLIP method in the RS domain. ReAttnCLIP\cite{niu2026reattnclip} proposes a re-defined attention mechanism for CLIP, improving dense feature quality for RS OVSS.

However, all the above methods rely heavily on the CLIP attention mechanism. Their proposed model still operates within the CLIP framework, which inevitably leads to only marginal improvements in segmentation performance. Although SegEarth-OV3\cite{li2025segearth} explores the application of SAM3 in RS OVSS, it merely adopts SAM3 in a straightforward manner without fully exploiting its potential. In this work, we present a training-free OVSS method, RSGPNet, which addresses RS OVSS challenges through geometric prompting.

\section{METHODS}
\label{sec: method}

This section will describe the proposed RSGPNet and explain why it is important for OVSS in the RS domain. This section is organized as follows: Section III-A describes our TCM, Section III-B introduces the GRP, and Section III-C describes the core principles of the CVM. 

\begin{figure*}[!th]
\centering
\includegraphics[width=7in]{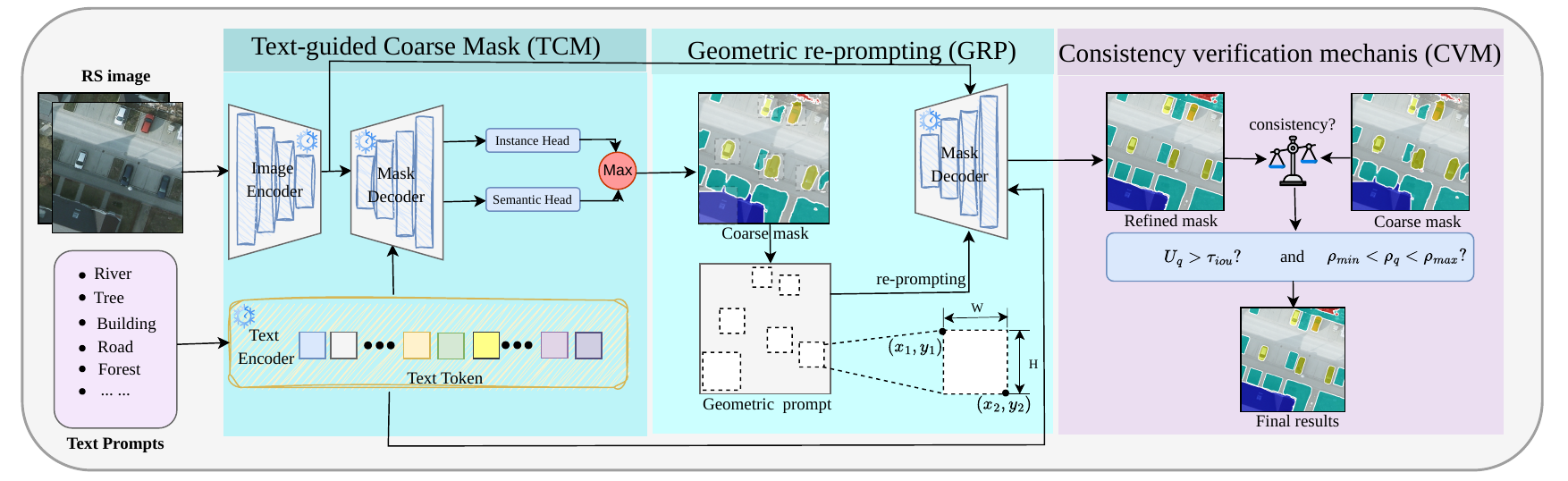}\vspace{-0.3cm}
\caption{Overall architecture of RSGPNet, featuring the Text-guided Coarse Mask module (TCM), Geometric Re-prompting Module (GRP), and Coarse-to-fine Consistency Verification Mechanism (CVM).}
\label{fig: framework}
\vspace{-0.5cm}
\end{figure*}

\subsection{Text-guided Coarse Mask}

Unlike conventional OVSS methods that generate the final segmentation result in a single step, RSGPNet performs segmentation in a coarse-to-fine manner through multiple modules. Specifically, the TCM module utilizes the input textual and visual prompts to generate a coarse mask with the help of a visual language model (VLM). The primary role is to provide a reliable region for the subsequent GRP module to generate geometric prompts. The detailed formulation process is described bellow. Given an input image $I \in \mathbb{R}^{H \times W \times 3}$ and a set of textual prompts $\mathbb{T}=\{T_1, T_2, \dots, T_Q\}$, the output of the VLM for the $q$-th prompt $T_q \in \mathbb{T}$ is formulated as follows:

\begin{equation}
\{M^{ins}_{q,k}\}^{K_q}_{k=1}, \{s^{obj}_{q,k}\}^{K_q}_{k=1}, M_q^{sem} = \text{VLM($I, T_q$)},  
\end{equation}
where $M^{ins}_{q,k}$ represents the mask logit of the $k$-th instance for the $q$-th prompt, $s^{obj}_{q,k}$ is its corresponding object score, and $M_q^{sem}$ denotes the resulting semantic mask. The logit outputs of the instance and semantic branches are defined as:
\begin{equation}
L_q^{ins}(x)= max(M^{ins}_{q,k}\ \cdot s^{obj}_{q,k}), 
\end{equation}
\begin{equation}
L_q^{sem} (x)=M_q^{sem}(x).
\end{equation}
Here x denotes the pixel location: 
\begin{equation}
x=(u,v),u\in\{1,2,...,H\},v\in\{1,2,....,W\}. 
\end{equation}
The original coarse segmentation logit can be expressed as:
\begin{equation}
L_q(x)=max(L_q^{sem} (x), L_q^{ins}(x)).
\end{equation}
Based on the unique characteristics of RS imagery, we execute semantic and instance segmentation independently using the VLM\cite{li2025segearth}. Specifically, instance segmentation is more effective for discrete, countable objects such as vehicles and buildings, whereas semantic segmentation yields superior performance on continuous regions like roads and rivers. Finally, the coarse segmentation logits are mapped to probabilities via an activation function to yield the initial coarse masks:
\begin{equation}
Y_q(x)=\mathbb I [\sigma(L_q(x))>\tau_m].
\end{equation}
where $\sigma$ denotes the sigmoid function, $\tau_m$ is the threshold for the coarse mask, and $\mathbb I$ represents the indicator function. TCM generates a coarse mask from the text prompt and input image, establishing the foundation for the geometric prompting in the following GRP module.

\subsection{Geometric Re-prompting}

In this section, we state the core motivation and technical justification for proposing the GRP module within the context of RS OVSS. While traditional OVSS paradigms typically rely on standard text-image alignment, most open-vocabulary frameworks achieve only marginal performance gains by modifying internal CLIP attention mechanisms. In this work, we are the first to propose the GRP, which introduces geometric prompts in addition to textual prompts, enabling a second-round prompting of the VLM for refined segmentation. 

We observe that geometric prompting is not suitable for all categories. Specifically, objects with relatively regular shapes, such as vehicles, cropland, and forests, can benefit from geometric guidance. In contrast, for elongated and spatially continuous categories such as rivers, geometric prompts may disrupt the overall structural consistency. Therefore, we design a gating function to determine whether geometric prompting should be applied as follow.
\begin{equation}
G_q=\mathbb I [c(q)\in  C_r] \cdot \mathbb I[Y_q>A_{min}],
\end{equation}
where $c(q)$ denotes the semantic category corresponding to the $q$-th prompt, and $C_r$ represents the set of categories eligible for prompting, e.g., $C_r=\{Building,Tree, Car\}$. $A_{\min}$ denotes the minimum area threshold for a connected component, ensuring that the GRP is applied selectively to categories suited for geometric prompting. In addition, isolated regions with negligible pixel counts are filtered out to suppress noise and mitigate interference during the segmentation process. When $G_q=1$, the coarse mask generated by the TCM is decomposed into a set of connected components:
\begin{equation}
Y_q =\bigcup_{j=1}^{N} \Omega_{q,j},
\end{equation}
where $\Omega_{q,j}$ represents the $j$-th connected component. We extract the bounding rectangle of each component to obtain its coordinates, which are then transformed into a normalized, center-based bounding box representation:
\begin{equation}
(x_1,y_1,x_2,y_2) =\text{BBox( $\Omega_{q,j}$)},
\end{equation}
\begin{equation}
B_{q,j} =(\frac{x_1+x_2}{2W},\frac{y_1+y_2}{2H},\frac{x_2-x_1}{W},\frac{y_2-y_1}{H}).
\end{equation}
These coordinates then serve as geometric prompts, which are fed back into the VLM to generate the refined segmentation logits.
\begin{equation}
L_q^r =\text{VLM($I, T_q, B_{q,j}$)}.
\end{equation}
Finally, the refined segmentation logits are mapped to a probability distribution via an activation function to yield the final refined mask, i.e.,
\begin{equation}
Y_q^r=\mathbb I [\sigma(L_q^r)>\tau_m].
\end{equation}
By pioneering the use of geometric prompting, the GRP module achieves superior, highly refined segmentation results for the OVSS task.

\subsection{Consistency verification mechanism}

Despite its advantages, the GRP may propagate errors by reinforcing baseline segmentation flaws. Therefore, we propose the CVM to quantify consistency between the coarse and refined masks. And the consistency metric is formulated as follows.
\begin{equation}
U_q = \frac{|Y_q^r \cap Y_q|}{|Y_q^r \cup Y_q|},
\end{equation}
\begin{equation}
\rho_q = \frac{|Y_q^r|}{|Y_q|}.
\end{equation}

The refined mask is accepted only when all of the following conditions are satisfied.
\begin{equation}
U_q > \tau_{iou},
\end{equation}
\begin{equation}
\rho_{min}<\rho_q< \rho_{max}.
\end{equation}

Mathematically, the CVM evaluates mask consistency using the area IoU and the relative area variation, preventing re-prompted predictions from excessively deviating from the original semantics in extreme cases. This criterion is grounded in the observation that IoU changes is small unless the prompting process drastically alters the segmentation of the original category. In most cases, the model only refines fine-grained local details, such as object boundaries, rather than modifying the overall semantic prediction. Consequently, this simple consistency constraint preserves the advantages of geometric prompting while mitigating catastrophic drifts from the original segmentation result. All the above hyperparameters will be validated and selected through the ablation studies presented in our experimental section.

This section presents the core architecture and innovations of RSGPNet. Our framework achieves progressive, coarse-to-fine segmentation by coordinating three key mechanisms: TCM, GRP, and CVM. Unlike conventional OVSS approaches that depend solely on text prompts, this work is the first to leverage geometric prompting to refine localized mask predictions. The empirical superiority of this method is validated through extensive evaluations in the subsequent sections.

\section{Experiment}
\label{sec: experiment}
To thoroughly validate the effectiveness of the proposed framework, we conduct extensive experiments on several RS semantic segmentation benchmarks. We first introduce the experimental datasets and evaluation metrics, followed by comprehensive quantitative and qualitative comparisons with SOTA OVSS methods. Finally, we perform ablation studies to analyze the contribution of each component in the proposed framework.

\subsection{Datasets and Evaluation Metrics}
To comprehensively evaluate the proposed method, we conduct experiments on four widely used RS semantic segmentation benchmarks, namely LoveDA, Potsdam, Vaihingen, and UDD5.

\begin{enumerate}
\item \textbf{LoveDA \cite{Loveda}:} LoveDA is a large-scale RS dataset designed for land-cover semantic segmentation. It contains 5,987 high-resolution images with a spatial resolution of 0.3 m, collected from three Chinese cities, namely Nanjing, Changzhou, and Wuhan. The dataset covers both urban and rural scenarios and provides annotations for 166,768 semantic objects across multiple land-cover categories.

\item \textbf{Potsdam \cite{ISPRS_Potsdam}:} Potsdam is a high-resolution urban RS benchmark released by ISPRS. It consists of 38 aerial image tiles of Potsdam, Germany, each with a spatial resolution of 5 cm and a size of $6000 \times 6000$ pixels. The dataset provides RGB and infrared imagery with pixel-level annotations for six urban land-cover classes. Following common practice, the original images are cropped into 5,472 patches of size $512 \times 512$.

\item \textbf{Vaihingen \cite{ISPRS_Vaihingen}:} Vaihingen is another widely used urban RS benchmark provided by ISPRS. It contains 33 orthorectified aerial image tiles with a spatial resolution of 5 cm, covering a small village area in Vaihingen, Germany. The dataset includes RGB and infrared imagery and contains diverse urban structures, such as detached houses and multi-story buildings. The original images are cropped into patches of size $512 \times 512$ for evaluation.

\item \textbf{UDD5 \cite{chen2018large}:} UDD5 is a drone-based RS dataset jointly developed by Peking University and Beijing University of Posts and Telecommunications. It contains high-resolution UAV images collected from diverse urban environments across multiple Chinese cities. The dataset is designed for urban semantic segmentation and provides fine-grained pixel-level annotations for five semantic categories.
\end{enumerate}

Since RS datasets often employ varied terminology to describe same semantic categories, prompt engineering is essential to ensure a fair evaluation in the OVSS setting. The complete prompt list is provided in Table \ref{tab_I}, where synonymous descriptions for a given category are enclosed in square brackets. Adhering to the official nomenclature of each dataset, we introduce supplementary synonyms only when necessary to maintain consistency with existing RS OVSS benchmarks. Within each bracket, the first prompt denotes the official class name provided by the dataset.

\begin{table}[!t]
\caption{The prompt classes of the RS datasets.}
\centering
\begin{tabular}{c | >{\centering\arraybackslash}m{6.8cm}}
    \toprule
    \textbf{Datasets} & \textbf{Prompt Classes Name}\\
    \midrule
     \multirow{2}{*}{LoveDA}   & Background, [Building, House], Road, Water\\
     ~& [Barren,Bareland,Soil], [Forest,Tree], Agricultural\\
    \midrule
    \multirow{2}{*}{Potsdam}    & Road, Impervious surfaces, Building, Tree,\\
    ~&Grass,  Car, Clutter\\
    \midrule
    \multirow{2}{*}{Vaihingen}  & Road, Impervious surfaces, Building, Tree,\\
    ~&Grass,  Car, Clutter\\
    \midrule
    UDD5  & Vegetation, Building, Road, Vehicle, Background\\
    \bottomrule
\end{tabular}%
\label{tab_I}
\end{table}

Following the evaluation protocol adopted in previous OVSS studies, we use mean Intersection-over-Union ($mIoU$), mean F1-score ($mF1$), and mean Boundary F1 score ($mBF1$) as the primary evaluation metric\cite{li2025segearth, 11502025}. These metrics can effectively reflect the overall segmentation performance and boundary-region segmentation performance of different models on RS images. Specifically, mIoU is calculated as the average IoU across all semantic categories and serves as a comprehensive measure of segmentation performance:
\begin{equation}
IoU_i = \frac{TP_i}{TP_i+FP_i+FN_i},
\end{equation}
\begin{equation}
mIoU=\frac{1}{N}\sum_{i=1}^{N} IoU_i,
\end{equation}
where $TP_i$, $FP_i$, and $FN_i$ denote the true positives, false positives, and false negatives of class $i$, respectively, and $N$ represents the total number of semantic classes. By averaging the IoU scores over all classes, mIoU assigns equal importance to each category, thereby reducing the bias toward dominant classes. Therefore, it is particularly suitable for fair evaluation in multi-class RS semantic segmentation tasks. In addition, we report the $mF1$ to further evaluate the segmentation quality of different methods. The $mF1$ is calculated as follows:
\begin{equation}
\label{deqn_ex1a}
R_i=\frac{TP_i}{TP_i+FN_i},
\end{equation}
\begin{equation}
\label{deqn_ex1a}
P_i=\frac{TP_i}{TP_i+FP_i},
\end{equation}
\begin{equation}
\label{deqn_ex1a}
F1_i=\frac{2 \times R_i \times P_i}{R_i+P_i}.
\end{equation}
\begin{equation}
mF1=\frac{1}{N}\sum_{i=1}^{N} F1_i,
\end{equation}
where $F1_i$ is the $F1$ score for class $i$. Additionally, we use $BF1$ as a metric to evaluate the accuracy of segmentation in class boundary regions. In our paper, we define the boundary distance as follows:
\begin{equation}
\label{deqn_ex1a}
Dist=Max(H, W) \times \epsilon.
\end{equation}
where $H$ and $W$ denote the height and width of the image, respectively, and $\epsilon$ is a predefined constant set to 0.02 in our experiments. The boundary distance threshold is defined as $Dist$, where pixels located within a distance smaller than $Dist$ from the object boundary are considered boundary pixels. The $mBF1$ is obtained by averaging the $BF1$ scores on all classes, ensuring equal contributions from each category and preventing the evaluation from being dominated by large classes.

\subsection{Comparison to SOTA}
We first compare our method with three SOTA RS OVSS models, i.e., OVRS \cite{10962188}, SegEarth-OV \cite{li2025segearth}, RSCLIP\cite{11502025}, ReAttnCLIP\cite{niu2026reattnclip} and SegEarth-OV3\cite{li2025segearth}. Moreover, we additionally select six SOTA OVSS models from natural images for comparison, including standard CLIP \cite{radford2021learning}, MaskCLIP \cite{zhou2022extract}, SCLIP \cite{wang2024sclip}, GEM \cite{bousselham2024grounding}, NACLIP\cite{hajimiri2025pay} and ClearCLIP \cite{lan2024clearclip}. These models are trained via publicly available source code and default configurations. We use their base versions, and some reported results are cited directly from their papers. It is worth noting that the OVRS model is a re-training method, and we compare it with the results reported in their paper using the iSAID dataset as the training set. The VLM model used in our RSGPNet is SAM3\cite{carion2025sam}.

\subsubsection*{\bf Quantitative results}

Table \ref{tab_main} presents the quantitative comparison of different methods. In general, OVSS approaches specifically designed for RS imagery achieve better performance on RS datasets than conventional OVSS methods. Specifically, compared with traditional models such as CLIP and MaskCLIP, RS-oriented methods including OVRS, SegEarth-OV, RSCLIP, and ReAttnCLIP demonstrate notable improvements. This can be attributed to their targeted adaptations to the unique characteristics of RS imagery, such as multi-scale object distributions and the importance of local attention mechanisms. However, these methods are all built upon CLIP mechanism and therefore has inherent limitations of the CLIP framework. SegEarth-OV3 adopts SAM3 as both the encoder and decoder, leading to a substantial performance gain over previous approaches. Our proposed RSGPNet achieves the highest mIoU scores on the Potsdam, LoveDA, and Vaihingen datasets, while ranking second on UDD5. Moreover, RSGPNet consistently obtains the best mF1 scores across all datasets. On average, RSGPNet surpasses the second-best method by 0.6\% in mIoU and 0.9\% in mF1, highlighting its superior generalization capability. These results demonstrate that the proposed method achieves SOTA performance across multiple RS benchmarks. Additionally, our RSGPNet achieves the highest mBF1 scores across all datasets, consistently outperforming previous methods. These results demonstrate that RSGPNet provides superior segmentation quality in local regions, particularly around object boundaries, indicating its stronger capability for preserving fine-grained structural details and accurately delineating class contours.

Additionally, we selected a subset of representative RS classes with different scales for further analysis, including those classes for which our method achieved the second-best IoU. The corresponding results are presented in Table \ref{tab: class}. It can be observed that RSGPNet consistently achieves the highest $BF1$ scores across all selected classes, even for classes where its IoU ranks second. This demonstrate its superior capability in preserving boundary details and improving segmentation quality around object contours.

\begin{table*}[!t]
\centering
\caption{Comparison results with SOTA methods on RS datasets. The best results are highlighted in bold.}
\label{tab_main}
\scriptsize
\setlength{\tabcolsep}{3pt}
\renewcommand{\arraystretch}{1.05}
\resizebox{\textwidth}{!}{
\begin{tabular}{l ccc ccc ccc ccc|ccc}
\toprule
\textbf{Models} 
& \multicolumn{3}{c}{\textbf{LoveDA}}
& \multicolumn{3}{c}{\textbf{Potsdam}}
& \multicolumn{3}{c}{\textbf{Vaihingen}}
& \multicolumn{3}{c}{\textbf{UDD5}}
& \multicolumn{3}{c}{\textbf{Average}} \\
\cline{2-16}
& mIoU & mF1 & mBF1
& mIoU & mF1 & mBF1
& mIoU & mF1 & mBF1
& mIoU & mF1 & mBF1
& mIoU & mF1 & mBF1 \\
\midrule

CLIP\cite{radford2021learning} 
& 12.4 & 22.4 & 4.9
& 14.5 & 25.7 & 6.7
& 10.3 & 19.6 & 3.8
& 9.5 & 17.7 & 2.6
& 11.7 & 21.4 & 4.5 \\

MaskCLIP\cite{zhou2022extract}
& 27.8 & 39.3 & 14.4
& 31.7 & 45.0 & 13.5
& 24.7 & 36.5 & 15.8
& 32.4 & 45.4 & 21.2
& 29.2 & 41.6 & 16.2 \\

SCLIP\cite{wang2024sclip}
& 30.4 & 44.8 & 24.2
& 36.6 & 50.4 & 20.1
& 28.4 & 40.1 & 18.9
& 38.7 & 52.6 & 26.8
& 33.5 & 47.0 & 22.5 \\

GEM\cite{bousselham2024grounding}
& 31.6 & 45.6 & 25.8
& 36.5 & 50.4 & 19.9
& 24.7 & 36.7 & 15.8
& 41.2 & 57.8 & 39.6
& 33.5 & 47.6 & 25.3 \\

NACLIP\cite{hajimiri2025pay}
& 32.3 & 45.7 & 26.6
& 40.1 & 55.9 & 22.3
& 26.1 & 38.2 & 17.1
& 42.6 & 58.6 & 42.5
& 35.3 & 49.6 & 27.1 \\

ClearCLIP\cite{lan2024clearclip}
& 32.4 & 46.7 & 26.1
& 40.9 & 56.0 & 22.8
& 27.3 & 39.9 & 17.7
& 41.8 & 58.5 & 40.8
& 35.6 & 50.3 & 26.9 \\

OVRS\cite{10962188}
& 30.8 & 44.8 & 24.1
& 27.4 & 39.6 & 12.0
& 20.8 & 33.5 & 12.6
& 30.2 & 43.5 & 18.6
& 25.4 & 40.4 & 16.8 \\

SegEarth-OV\cite{li2025segearth}
& 36.9 & 50.9 & 28.2
& 47.1 & 62.8 & 25.2
& 29.1 & 43.3 & 19.5
& 50.6 & 65.1 & 54.3
& 40.9 & 55.5 & 31.8 \\

RSCLIP\cite{11502025}
& 38.0 & 51.9 & 29.2
& 47.4 & 64.0 & 25.9
& 28.9 & 43.9 & 19.6
& 50.8 & 66.2 & 54.7
& 41.3 & 56.5 & 32.4 \\

ReAttnCLIP\cite{niu2026reattnclip}
& 37.0 & 51.8 & 28.9
& 48.7 & 65.4 & 24.6
& 29.9 & 45.2 & 19.3
& 53.7 & 67.4 & 51.9
& 42.2 & 57.5 & 31.2 \\

SegEarth-OV3\cite{li2025segearth}
& 47.4 & 62.8 & 54.3
& 57.8 & 70.2 & 52.1
& 60.8 & 71.1 & 57.2
& \textbf{71.0} & 81.6 & 77.0
& 59.3 & 71.4 & 60.2 \\

\textbf{RSGPNet (Ours)}
& \textbf{47.8} & \textbf{64.0} & \textbf{56.2}
& \textbf{59.7} & \textbf{71.7} & \textbf{57.2}
& \textbf{61.1} & \textbf{71.5} & \textbf{59.8}
& \textbf{71.0} & \textbf{81.8} & \textbf{78.1}
& \textbf{59.9} & \textbf{72.3} & \textbf{62.8} \\

\bottomrule
\end{tabular}
}
\end{table*}

\begin{table*}[!t]

    \caption{Comparison Results with SOTA Methods on Different-Scale Categories in RS Datasets.}
    \label{tab: class} 
    \centering
    \normalsize
    \begin{tabular*}{\textwidth}{@{\extracolsep{\fill}} c c c c c c c @{\hspace{0.5em}} c @{\hspace{0.5em}} c @{\hspace{0.5em}}  c c @{\hspace{0.5em}}}
        \toprule
        \textbf{Models} &\multicolumn{2}{c} {\textbf{Car}}  &\multicolumn{2}{c} {\textbf{Vehicle}}
        &\multicolumn{2}{c} {\textbf{Tree} }
        &\multicolumn{2}{c} {\textbf{Agricultural}}
         &\multicolumn{2}{c} {\textbf{Building} }\\
        \cline{2-3} \cline{4-5} \cline{6-7} \cline{8-9}\cline{10-11}
        ~&IoU&BF1  &IoU&BF1 &IoU&BF1 &IoU&BF1  &IoU&BF1 
        \\
        \midrule
        CLIP\cite{radford2021learning} & 15.6 & 5.3 & 36.2& 25.1 & 19.9& 6.5 &18.7 & 7.1 & 24.1& 13.4\\
        MaskCLIP\cite{zhou2022extract} & 28.8 & 8.9 & 62.4 & 46.2 
         & 34.8& 10.8 & 33.8 & 12.0 & 41.0& 21.3\\
         SCLIP\cite{wang2024sclip} & 30.4 & 10.0 & 72.5 & 52.2
         & 37.8& 12.1 &36.2 & 13.9& 42.1& 21.7\\
        GEM\cite{bousselham2024grounding} & 30.9 & 11.8 & 69.0 & 51.8 & 36.9 & 11.8& 39.9& 14.2 & 44.1 & 22.3 \\
        NACLIP\cite{hajimiri2025pay} & 34.2 & 12.6 & 71.9  & 53.6  & 38.6  & 12.5 & 41.0 & 15.1 & 46.8 & 23.0 \\
        ClearCLIP\cite{lan2024clearclip} & 34.5 & 12.9 & 71.8 & 53.5 & 39.8 & 12.7 & 41.2 & 15.4 & 47.4 & 22.9 \\
        OVRS\cite{10962188} & 26.7 & 8.7 & 60.6 & 44.2 & 33.5 & 10.5  & 30.7 & 11.9 & 40.1 & 20.8 \\
        SegEarth-OV\cite{li2025segearth} & 42.1 & 15.9 & 78.3 & 60.2 & 50.1 & 17.9 & 49.1 & 21.2 & 55.3 & 27.9 \\
        RSCLIP\cite{11502025}  & {44.1}& {17.6} & \textbf{80.5} & {62.5} & {53.8} & {19.6} &\textbf{52.0}& {22.6} &{58.5}&{29.2}\\
        ReAttnCLIP\cite{niu2026reattnclip}  & 44.6 & 18.0 & 75.4 & 62.2 & 46.8 & 15.7 & 46.1 & 18.8  & 57.7 & 24.5\\
        SegEarth-OV3\cite{li2025segearth}  & 76.7 & 79.3 & 67.2  & 75.4 & 72.3 & 64.5 & 46.8 & 36.2 & 78.6 & 60.3 \\
        \textbf{RSGPNet (Ours)}  & \textbf{78.9} & \textbf{84.0} & 69.7 & \textbf{78.9} & \textbf{74.0} & \textbf{69.1} &47.8 & \textbf{39.6} & \textbf{84.7} & \textbf{66.7}\\
        \bottomrule
    \end{tabular*}

\end{table*}

\subsubsection*{\bf Visualization Comparison}
\begin{figure*}[!t]
\centering
\includegraphics[width=\textwidth]{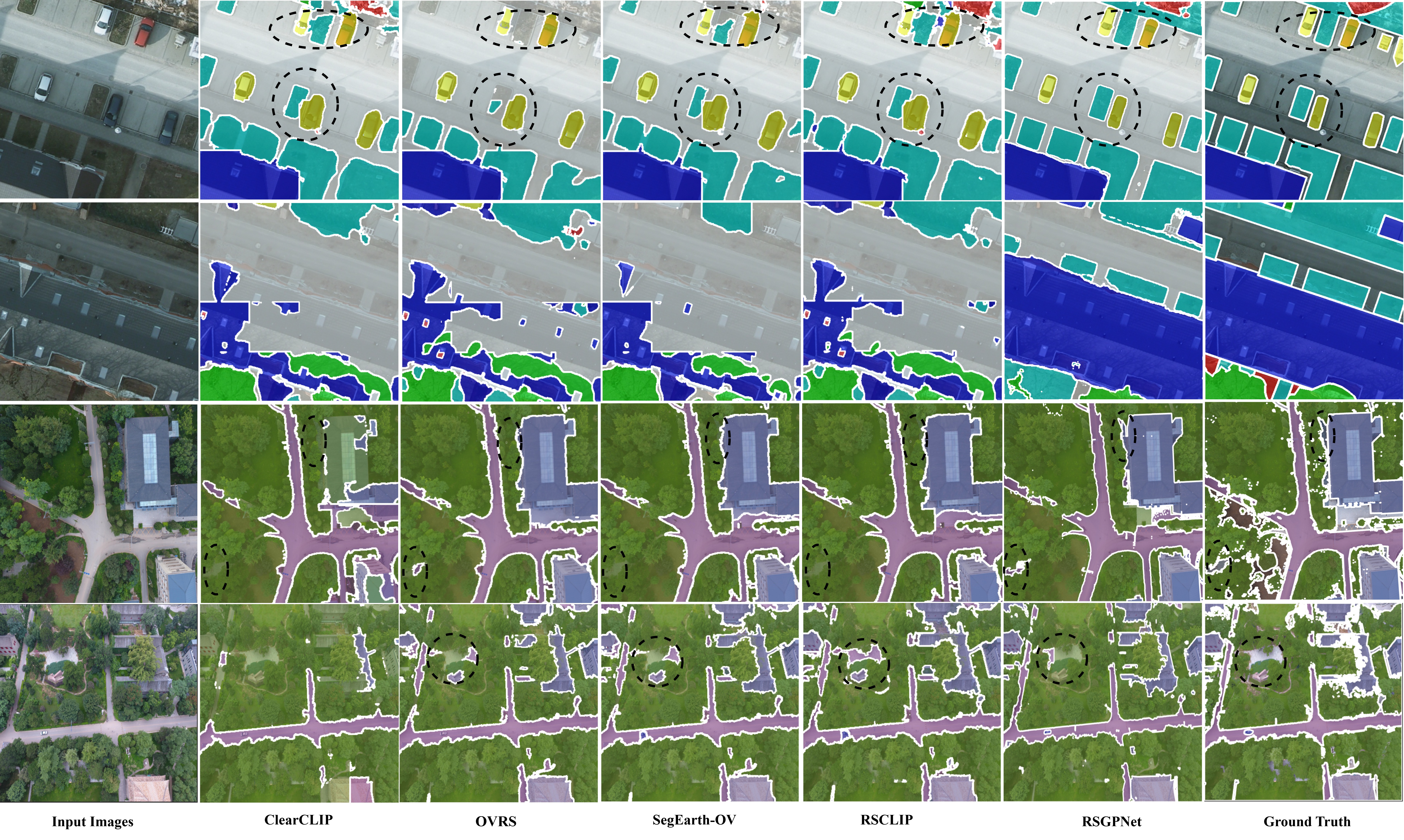}
\vspace{-0.5cm}
\caption{‌Visualization Comparison of SOTA OVSS Methods on RS Datasets. From left to right: Input Image, ClearCLIP, OVRS, SegEarth-OV, RSCLIP, our RSGPNet and Ground Truth.}
\vspace{-0.5cm}
\label{fig: vi}
\end{figure*}

We present the qualitative segmentation results of RSGPNet and compare them with those of other SOTA methods as shown in Fig.\ref{fig: vi}. The semantic segmentation results shown in the figure are multi-class. To present the results more intuitively, we overlaid the segmentation masks on the original images.

The results demonstrate that all competing methods suffer from severe segmentation artifacts on challenging scenes, as exemplified by the image in the second row. Specifically, the visual appearance of the building regions closely mimics the surrounding grassland, causing existing models to fail catastrophically. In contrast, our framework leverages the proposed geometric re-prompting mechanism to execute a secondary, fine-grained assessment of critical ambiguous regions, thereby achieving significantly superior segmentation accuracy.

For some RS images, such as those shown in the first, third, and fourth rows, previous methods appear to achieve satisfactory segmentation results overall. However, they still fail to capture fine-grained details accurately. Specifically, for the image in the first row, previous methods produce segmentation errors in the regions marked by the black dashed ellipses, where the categories car and low vegetation are incorrectly classified. For the image in the third row, segmentation errors occur in the building and road regions highlighted by the black dashed ellipses. Furthermore, for the bare-land area in the fourth-row image, previous methods exhibit large-scale misclassification labeling the region as building. In contrast, our method yields the smallest misclassified area among all compared approaches. The visualization comparison results demonstrate that our method achieves superior qualitative segmentation performance.

\subsubsection*{\bf Attention Visualization Comparison}

In this subsection, attention visualization was employed to compare the attention distributions of different models across various regions. 

\begin{figure*}[!t]
\centering
\includegraphics[width=\textwidth]{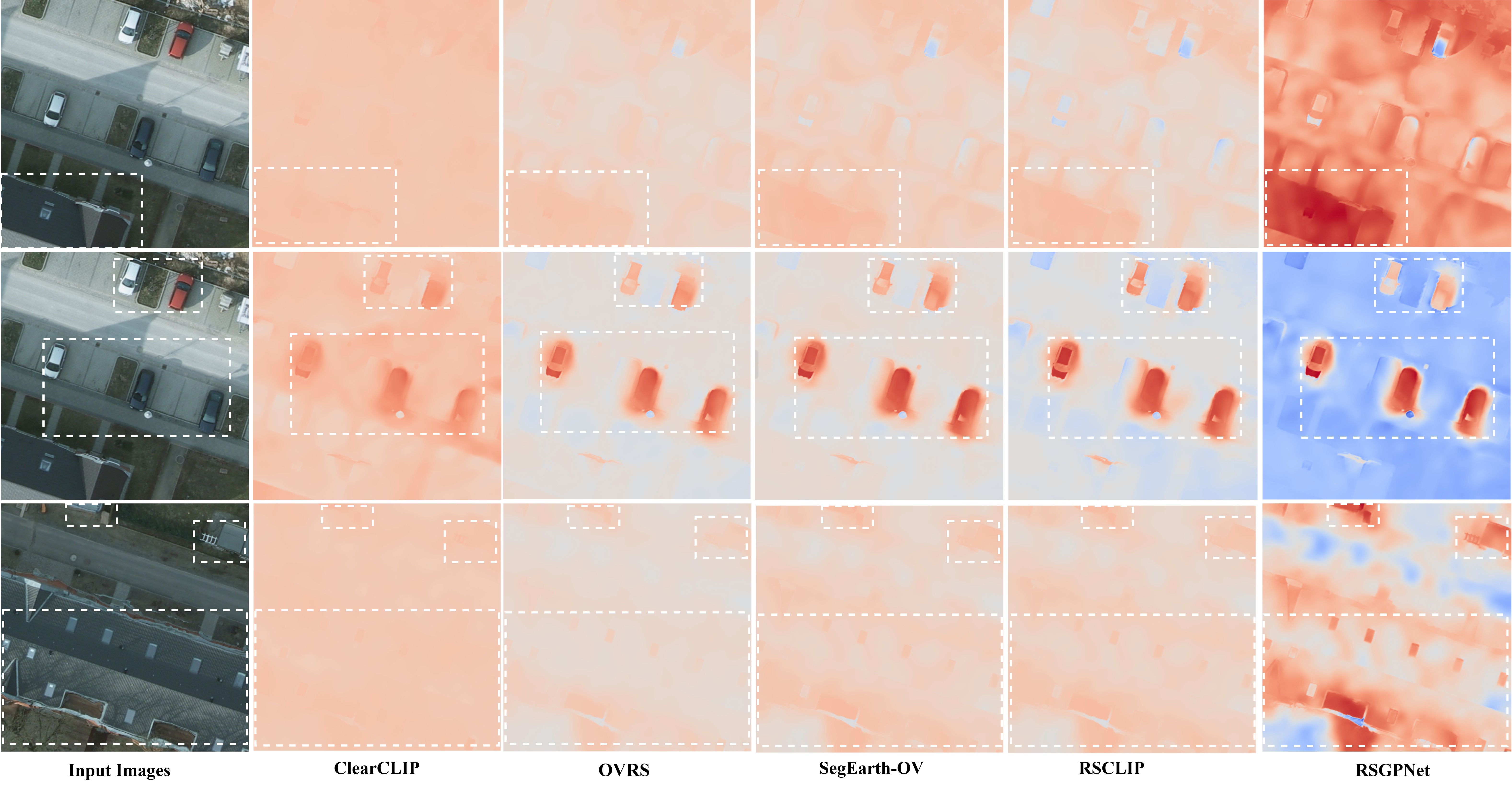}
\vspace{-0.5cm}
\caption{Attention‌ visualization Comparison of SOTA OVSS Methods on RS Datasets.}
\vspace{-0.5cm}
\label{fig: attention}
\end{figure*}

The visualization results are presented in Fig. \ref{fig: attention}. The first row displays the attention maps of the competing models for the building structure situated in the lower-left corner, where warm tones indicate higher activation levels. As illustrated, ClearCLIP exhibits the weakest attention, characterized by an almost uniform distribution across the entire scene. OVRS, SegEarth-OV, and RSCLIP show slightly stronger responses; however, their responses are still not clearly distinguishable from those of the surrounding regions. In contrast, our proposed RSGPNet is able to effectively isolates the building area from other regions, demonstrating a superior capability to focus on the target object. Similarly, for the building region shown in the third row, none of the other models are able to correctly focus on the relevant area, whereas our proposed RSGPNet successfully concentrates its attention on the target object.

For the car category in the second row, although most models correctly attend to the target region, their activation intensities are different. Specifically, ClearCLIP exhibits the lowest attention response, whereas OVRS, SegEarth-OV, and RSCLIP yield relatively stronger activations. However, none of these baseline models can clearly decouple the target car from adjacent semantic classes. In contrast, our proposed RSGPNet produces highly activated responses (shown in deep red) over the car regions, while assigning significantly lower responses (shown in deep blue) to other categories. This demonstrates that RSGPNet not only strongly emphasizes the target class but also suppresses attention to irrelevant categories, yielding a highly discriminative and category-specific attention distribution.
\subsection{Ablation Study and Analysis}
To evaluate the efficacy of the proposed RSGPNet, we conduct a series of ablation experiments across the aforementioned benchmarks to systematically isolate the contributions of each individual module.
\subsubsection*{\bf Ablation of each component}

We first analyze the individual and joint effects of the TCM, GRP, and CVM. Notably, these architectural components exhibit inherent functional dependencies. For instance, GRP relies on the coarse masks generated by the TCM to construct geometric prompts; thus, the TCM and GRP are invariably ablated jointly. Similarly, because the CVM is explicitly designed to evaluate consistency between coarse and refined masks, its isolation is non-viable without both inputs. Consequently, the CVM is exclusively evaluated when its prerequisite mask generation modules are active. Unless otherwise specified, $mIoU$ serves as the primary evaluation metric throughout this section.

\begin{table*}[!t]
\caption{ Ablation study for different components of our RSGPNet. \label{tab:ablation}}
\centering
\small
\begin{tabular*}{\textwidth}{@{\extracolsep{\fill}} c |c c @{\hspace{0.5em}} c @{\hspace{0.75em}} c  @{\hspace{0.5em}} | c  @{\hspace{0.5em}} c  @{\hspace{0.5em}} c  @{\hspace{0.5em}}   c  @{\hspace{0.5em}} | c  @{\hspace{0.5em}}} 
		\toprule
		~ &\textbf{Baseline}&\textbf{TCM}&\textbf{GRP}&\textbf{CVM}  &\textbf{LoveDA} &\textbf{Potsdam}  &\textbf{Vaihingen} &\textbf{UDD5}&\textbf{Average}\\
		\midrule
		A & $\checkmark$ & ×  & ×  & ×  & 47.4 & 57.8 & 60.8 & 71.0 & 59.3\\
        B & $\checkmark$ & $\checkmark$  & $\checkmark$  & ×& 47.6 & 59.0  & 60.9  & 71.0 & 59.6 \\
        C & $\checkmark$ & $\checkmark$  & $\checkmark$  & $\checkmark$ & \textbf{47.8} & \textbf{59.7}  & \textbf{61.1}  & \textbf{71.0} & \textbf{59.9} \\
		\bottomrule
	\end{tabular*}
\end{table*}

The experimental results are summarized in Table \ref{tab:ablation}. The results indicate that incorporating the TCM and GRP modules improves the model performance. Furthermore, the best performance is achieved after applying the consistency computation through the CVM module. It is worth noting that our model architecture differs fundamentally from conventional OVSS models. As a result, even the baseline significantly outperforms the previous SOTA RS OVSS method, ReAttnCLIP\cite{niu2026reattnclip}, by more than ten percentage points. Therefore, each performance gain observed in the ablation study is particularly noteworthy.
The $\text{mIoU}$ on the UDD5 dataset exhibits a negligible numerical shift primarily because the benchmark's smaller scale caps absolute metric fluctuations, yielding only marginal fractional gains that are obscured when results are rounded to one decimal place. Nonetheless, the consistent performance gains across all other RS benchmarks robustly validate both the efficacy and generalizability of our proposed framework.

\begin{table}[h]
\caption{$\tau_m$ parameter study for TCM.}
\label{tab: TCM}
\centering
\begin{tabularx}{\linewidth}{c|X X X X|c}
    \toprule
    \textbf{$\tau_m$}&\textbf{LoveDA}  & \textbf{Potsdam}&\textbf{Vaihingen}&\textbf{UDD5}&\textbf{Average} \\
    \midrule
    0.5 & 47.7 & 59.6  & 61.0  & 69.9 & 59.6   \\
    \midrule
    0.6 & \textbf{47.8} & \textbf{59.7}  & \textbf{61.1}  & \textbf{71.0} & \textbf{59.9}   \\
    \midrule
    0.7 & 47.7 & 59.6  & 61.0  & 71.0 & 59.8   \\
      \midrule
    0.8 & 47.6 & 59.5  & 60.8  & 69.9 & 59.5  \\
    \bottomrule
\end{tabularx}
\end{table}
\subsubsection*{\bf Ablation study for TCM}
We conducted an ablation study to investigate the impact of the coarse-mask threshold $\tau_m$ on model performance, selecting four representative values: 0.5, 0.6, 0.7, and 0.8. As shown in Table \ref{tab: TCM}, the effect of $\tau_m$ on the results is relatively minor, with the highest $IoU$ achieved at $\tau_m=0.6$. This peak occurs because an excessively low threshold provides insufficient discriminability, whereas a threshold that is too high filters out too many coarse masks.

\begin{table}[h]
\caption{$A_{min}$ parameter study for GRP.}
\label{tab: GRP1}
\centering
\begin{tabularx}{\linewidth}{c|X X X X|c}
    \toprule
    \textbf{$A_{min}$}&\textbf{LoveDA}  & \textbf{Potsdam}&\textbf{Vaihingen}&\textbf{UDD5}&\textbf{Average} \\
    \midrule
    60 & 47.7 & 59.6  & 60.9  & 69.8 & 59.5   \\
    \midrule
    70 & 47.7 & 59.7  & 61.0  & 69.9 & 59.6   \\
    \midrule
    80 & \textbf{47.8} & \textbf{59.7}  & \textbf{61.1}  & \textbf{71.0} & \textbf{59.9}   \\
    \midrule
    90 & 47.7 & 59.6  & 61.0  & 71.0 & 59.8   \\
    \midrule
    100 & 47.5 & 59.4  & 60.9  & 69.8 & 59.4  \\
    \bottomrule
\end{tabularx}
\end{table}
\subsubsection*{\bf Ablation study for GRP}
We further analyze the impact of the minimum connected-component area threshold ($A_{min}$) on model performance. As shown in Table \ref{tab: GRP1}, varying $A_{min}$ yields different segmentation results. The experimental results indicate that $A_{min}$ influences IoU, although the impact is not highly significant. Specifically, as $A_{min}$ increases, the IoU first improves and then declines, peaking at approximately 80 pixels. This trend can be attributed to the fact that a very small $A_{min}$ causes geometric prompting to be applied to almost all regions, potentially introducing noise. Conversely, an excessively large $A_{min}$ prevents small objects, such as cars, from benefiting from geometric enhancement. Based on these observations, $A_{min}=80$ is adopted as the final setting in this work.

Furthermore, taking the Potsdam dataset as an example, we analyze the impact of $C_r$, which denotes the set of categories suitable for geometric prompting, on IoU performance. The experimental results are presented in Table \ref{tab: GRP2}. When $C_r$ contains only the Impervious Surfaces, Low Vegetation, and Background categories, the IoU remains almost unchanged or even decreases slightly, indicating that these categories are unsuitable for the GRP module. In contrast, when $C_r$ includes other categories, improvements in IoU are observed. The best category combination achieves a 1.9\% IoU gain over the baseline. These results validate the effectiveness of the GRP module.
\begin{table}[h]
\caption{Ablation study of $C_r$ for geometric prompting.}
\label{tab: GRP2}
\centering
\begin{tabularx}{\linewidth}{c|X X X X X X|c}
    \toprule
    ~ & \textbf{Imp.}&\textbf{Build.}&\textbf{Tree} &\textbf{Low.} &\textbf{Car} &\textbf{Back.} &\textbf{mIoU}\\
    \midrule
    A   & ~ & ~ & ~ & ~& ~& ~& 57.8 \\
    \midrule
    B  & $\checkmark$ & ~ & ~& ~& ~& &57.8  \\ 
   \midrule 
    C  & ~ & $\checkmark$ & ~ & ~ & ~ & ~ &58.1 \\
    \midrule
    D  & ~ & ~ & $\checkmark$ & ~ & ~ & ~ &59.3 \\
    \midrule
    E  & ~ & ~ & ~ & $\checkmark$ & ~ & ~ &57.8 \\
    \midrule
    F   & ~ & ~ & ~& ~ & $\checkmark$ & &59.0  \\
    \midrule
    G  & ~ & ~ & ~& ~ & ~& $\checkmark$ & 57.7 \\ 
    \midrule
    H  & ~ & $\checkmark$ & $\checkmark$  & ~ & ~ & & 59.6  \\
    \midrule
    I  & ~ & $\checkmark$ & $\checkmark$  & ~& $\checkmark$ & & \textbf{59.7}  \\ 
    \bottomrule
\end{tabularx}
\end{table}

\subsubsection*{\bf Ablation study for CVM}
\begin{figure}[h]
\centering
\includegraphics[width=3.5in]{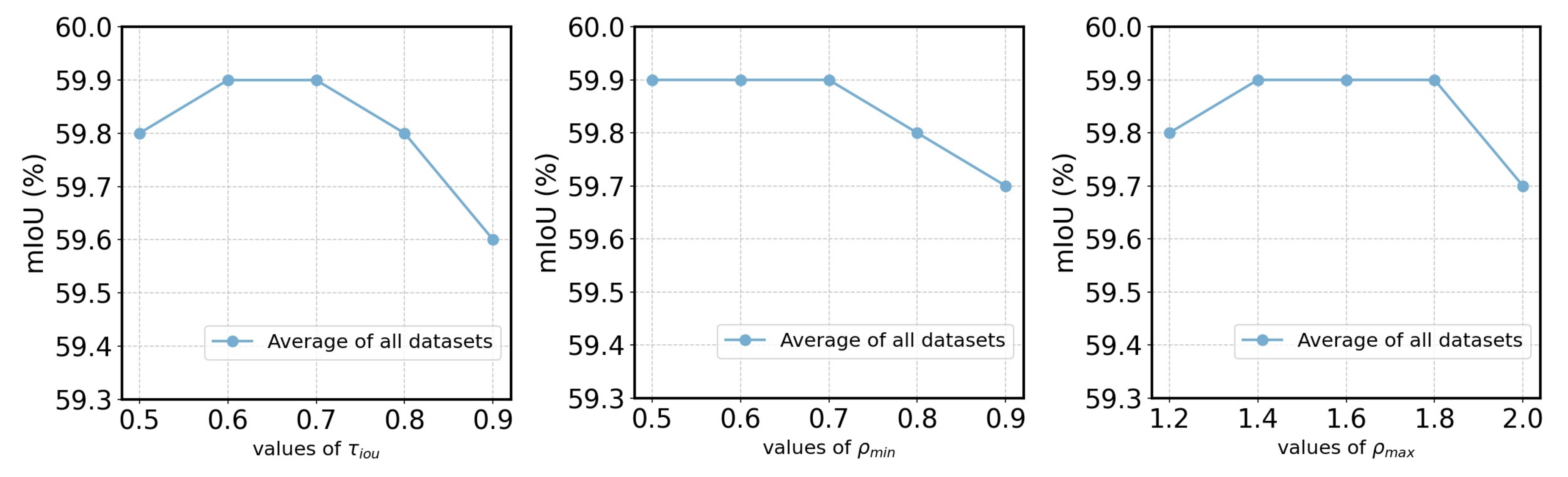}
\vspace{-0.5cm}
\caption{The ablation study of the hyperparameters $\tau_{iou}$, $\rho_{min}$, and $\rho_{max}$ for CVM}
\label{fig: ab_CVM}
\vspace{-0.5cm}
\end{figure}

We investigate the impact of the hyperparameters $\tau_{iou}$, $\rho_{min}$, and $\rho_{max}$ on the performance of the proposed model. Specifically, we compute the average mIoU across all RS datasets for different values of $\tau_{iou}$, $\rho_{min}$, and $\rho_{max}$, and the corresponding results are shown in Fig. \ref{fig: ab_CVM}.

The experimental results indicate that the mIoU initially increases and then decreases with increasing $\tau_{iou}$. This phenomenon can be attributed to the fact that a small $\tau_{iou}$ is insufficient to enforce consistency filtering, while an overly large $\tau_{iou}$ removes an excessive number of refined masks generated by GRP. Since $\rho_{min}$ and $\rho_{max}$ cannot be evaluated simultaneously, we first fixed $\rho_{max}$ at its initial value of 1.5 when investigating the effect of $\rho_{min}$. After determining the impact of $\rho_{min}$, we fixed $\rho_{min}$ and then evaluated different values of $\rho_{max}$. As shown in Fig. \ref{fig: ab_CVM}, the performance is not particularly sensitive to either $\rho_{min}$ or $\rho_{max}$. Good results are achieved when $\rho_{min}$ ranges from 0.5 to 0.7 and $\rho_{max}$ ranges from 1.4 to 1.8. Based on these observations, we finally set $\tau_{iou}$, $\rho_{min}$ and $\rho_{max}$ to 0.5, 0.5 and 1.5, respectively.

\section{Conclusion}
\label{sec: Conclusion}

We propose RSGPNet, a novel training-free OVSS method for RS images. Existing OVSS approaches primarily focus on improving the CLIP attention mechanism, yet they remain constrained by the inherent limitations of the CLIP architecture itself. This purely image-text matching paradigm causes OVSS methods to still face significant limitations in their ability to accurately segment local regions in RS images. To address this issue, RSGPNet is the first to introduce Geometric prompting, a new paradigm that leverages geometric prompts to achieve more accurate localization and segmentation of local regions in RS imagery. Specifically, RSGPNet consists of three key components that work collaboratively: TCM, GRP, and CVM. TCM utilizes text prompts and the input image to construct coarse segmentation masks. GRP converts these coarse masks into geometric box prompts and feeds them back into the segmentation model to generate refined masks. CVM employs consistency verification to prevent the prompting process from reinforcing erroneous regions and propagating segmentation errors.

Extensive open-vocabulary experiments on four public RS datasets demonstrate that RSGPNet achieves SOTA performance, strongly validating its effectiveness in RS vision tasks. Furthermore, our method produces visually significant improvements in both qualitative segmentation results and attention map visualizations, providing compelling evidence for the effectiveness of geometric prompting. This work offers a novel perspective for advancing research in RS OVSS.


\bibliographystyle{IEEEtran}
\bibliography{ref}

\end{document}